\documentclass[10pt, a4paper]{article}

\usepackage[final]{lrec2026} 

\usepackage[edges]{forest}
\usepackage{ccg}
\usepackage{amstext}
\usepackage{subcaption}  
\usepackage{booktabs}
\usepackage{algpseudocode}
\usepackage{adjustbox}
\usepackage{multicol}
\usepackage{algorithm}

\usepackage{booktabs}
\usepackage{tabularx}
\usepackage{xcolor}
\usepackage{caption}
\usepackage{amsmath}
\usepackage{subcaption}
\usepackage{multirow}
\usepackage{cleveref}
\usepackage{bm}
\Crefname{section}{\S}{\S\S}

\usepackage{amssymb}
\usepackage[dvipsnames]{xcolor}
\usepackage{hyperref}
\usepackage{enumitem}

\newcommand{\hidden}[1]{}
\def\1{\bm{1}}

\usepackage{graphicx}

\title{What Kind of Language is Easy to Language-Model \\ Under Curriculum Learning?}

\name{Nadine El-Naggar, Tatsuki Kuribayashi, Ted Briscoe} 

\address{Mohamed bin Zayed University of Artificial Intelligence \\
         \{nadine.naggar, tatsuki.kuribayashi, ted.briscoe\}@mbzuai.ac.ae\\}

\abstract{
Many of the thousands of attested languages share common configurations of features, creating a spectrum from typologically very rare  (e.g., object-verb-subject word order) or impossible languages to very common combinations of features (e.g., subject-object-verb word order).
One central question is under what conditions such typological tendencies can be predicted, and specifically whether the learning bias of language models (LMs) is sufficient to reproduce such patterns.
In this study, we add one dimensionality to such analysis --- the learning scenario for LMs --- to explore its interaction with the inductive bias of LMs.
Specifically, as a first study, we examine the effect of curriculum learning (CL), as a developmentally motivated learning scenario, i.e., starting with simpler sentences rather than randomly-ordered input. 
We expand existing LM-based exploration~\cite{el2025gcg,elnaggar2025wordordersfacilitatelength} with a simple CL variant and find that CL substantially impacts the apparent inductive bias of LMs. 
 \\ \newline  }

\begin{document}

\maketitleabstract

\section{Introduction}

Natural languages (NLs) exhibit a range of properties, including different word order configurations, raising the question: which, if any, types of languages are easier for language models (LMs) to learn~\cite{cotterell+al.naacl18a,DBLP:conf/acl/MielkeCGRE19,DBLP:conf/acl/WhiteC20,borenstein-etal-2024-languages,arnett-bergen-2025-language}?
Related questions include why some grammatical feature combinations are typologically common while others are rare~\cite{wals}, and to what extent LMs' domain-general learning biases replicate these linguistic tendencies~\cite{chomsky2023noam,xu2025languagemodelslearntypologically}.
Existing studies have reported that, for example, the inductive bias of a specific class of LMs aligns with typological commonality, and the properties of such typologically-aligned models (e.g., working memory limitations) could partially explain such typological tendencies~\cite{DBLP:conf/acl/KuribayashiUYOB24,el2025gcg,elnaggar2025wordordersfacilitatelength}.
Although such computational simulations of language learnability can serve as proof-of-concept for the idea that learning biases shaping language~\cite{Culbertson2012-bo}, LMs' and humans' language acquisition settings are generally different (e.g., the amount of data), and to make such simulations more relevant to human language science, it is necessary to align LM training scenarios with humans' ones~\cite{Warstadt2022-kd}.

One factor that is relevant to human language acquisition and overlooked in existing LM-based simulations of typological patterns is the order of data presentation.
For example, children with severe working memory limitations may take in relatively simple sentences first through language acquisition~\cite{Hudson_Kam2005-xt,Kam2009-jz}. 
Such effects have typically been simulated with curriculum learning (CL) in computational simulations, which present training data from simpler to more complex sentences.

In this study, we explore one basic CL setting in the LM-based simulation of typological patterns. 
That is, our research question is \textit{what kind of language is easier for LMs to learn under CL}.
As an initial foray, we adopt a simple length-based CL~\cite{spitkovsky2009baby} and replicate existing studies~\cite{el2025gcg,elnaggar2025wordordersfacilitatelength} to analyze interaction effect between CL and LMs' inductive biases over diverse word orders.
Our experimental results demonstrate that the word-order preferences indeed change with the CL setting, and, somewhat surprisingly, the CL-based results deviate more from typological commonality, which raises several implications (\cref{sec:discussion}).

\begin{figure*}[t]
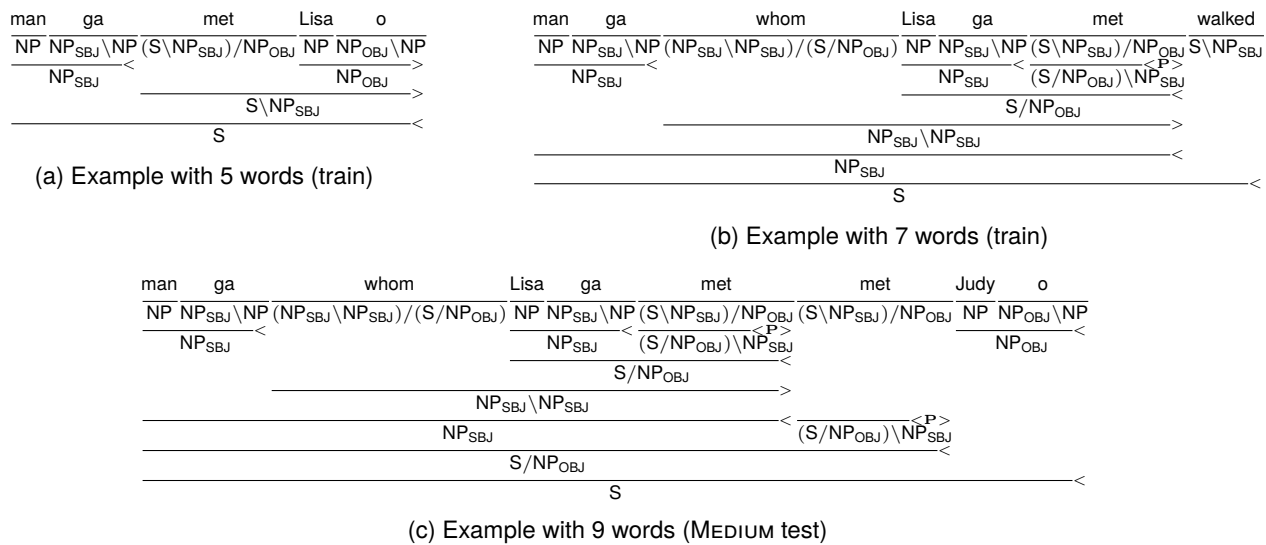

\scriptsize
\setlength{\arraycolsep}{0.5pt}
\centering

\renewcommand{\arraystretch}{0.5}

\begin{subfigure}[t]{0.32\textwidth}
\centering
\deriv{9}{
\text{man} & \text{ga} & \text{met} & \text{Lisa} & \text{o} \\
\uline{1} & \uline{1} & \uline{1} & \uline{1} & \uline{1} \\
\text{NP} & \text{NP$_\text{SBJ}$\bs NP} & (\text{S\bs NP$_\text{SBJ}$})/\text{NP$_\text{OBJ}$} & \text{NP} & \text{NP$_\text{OBJ}$\bs NP} \\
\bapply{2} & &  \fapply{2} \\ 
\mc{2}{\text{NP$_\text{SBJ}$}} & & \mc{2}{\text{NP$_\text{OBJ}$}}\\
& & \fapply{3} \\
& & \mc{3}{\text{S\bs NP$_\text{SBJ}$}} \\
\bapply{5} \\
\mc{5}{\text{S}}
}
\caption{Example with 5 words (train)}
\end{subfigure}
\hfill
\begin{subfigure}[t]{0.57\textwidth}
\centering
\deriv{16}{
\text{man} & \text{ga} & \text{whom} & \text{Lisa} & \text{ga} & \text{met} & \text{walked}\\
\uline{1} & \uline{1} & \uline{1} & \uline{1} & \uline{1} & \uline{1} & \uline{1}\\
\text{NP} & \text{NP$_\text{SBJ}$\bs NP} &
(\text{NP$_\text{SBJ}$\bs NP$_\text{SBJ}$})/(\text{S}/\text{NP$_\text{OBJ}$}) &
\text{NP} & \text{NP$_\text{SBJ}$\bs NP} &
(\text{S\bs NP$_\text{SBJ}$})/\text{NP$_\text{OBJ}$} &
\text{S\bs NP$_\text{SBJ}$}\\
\bapply{2} & & \bapply{2} & \permute{1} \\
\mc{2}{\text{NP$_\text{SBJ}$}} & & \mc{2}{\text{NP$_\text{SBJ}$}}  & \mc{1}{(\text{S}/\text{NP$_\text{OBJ}$})\bs\text{NP$_\text{SBJ}$}} & \\
& & & \bapply{3}\\
& & & \mc{3}{\text{S}/\text{NP$_\text{OBJ}$}}\\
& & \fapply{4}\\
& & \mc{4}{\text{NP$_\text{SBJ}$\bs NP$_\text{SBJ}$}}\\
\bapply{6}\\
\mc{6}{\text{NP$_\text{SBJ}$}}\\
\bapply{7}\\
\mc{7}{\text{S}}
}
\caption{Example with 7 words (train)}
\end{subfigure}
\\
\vspace{\baselineskip}
\begin{subfigure}[t]{0.8\textwidth}
\centering
\deriv{18}{
\text{man} & \text{ga} & \text{whom} & \text{Lisa} & \text{ga} & \text{met} & \text{met} & \text{Judy} & \text{o} \\
\uline{1} & \uline{1} & \uline{1} & \uline{1} & \uline{1} & \uline{1} & \uline{1} & \uline{1} & \uline{1}\\
\text{NP} & \text{NP$_\text{SBJ}$\bs NP} &
(\text{NP$_\text{SBJ}$\bs NP$_\text{SBJ}$})/(\text{S}/\text{NP$_\text{OBJ}$}) &
\text{NP} & \text{NP$_\text{SBJ}$\bs NP} &
(\text{S\bs NP$_\text{SBJ}$})/\text{NP$_\text{OBJ}$} &
(\text{S\bs NP$_\text{SBJ}$})/\text{NP$_\text{OBJ}$} &
\text{NP} & \text{NP$_\text{OBJ}$\bs \text{NP}}\\
\bapply{2} & & \bapply{2} & \permute{1} & & \bapply{2} \\
\mc{2}{\text{NP$_\text{SBJ}$}} & & \mc{2}{\text{NP$_\text{SBJ}$}} & \mc{1}{(\text{S}/\text{NP$_\text{OBJ}$})\bs\text{NP$_\text{SBJ}$}} & &   \mc{2}{\text{NP$_\text{OBJ}$}} \\
& & & \bapply{3}\\
& & & \mc{3}{\text{S}/\text{NP$_\text{OBJ}$}}\\
& & \fapply{4}\\
& & \mc{4}{\text{NP$_\text{SBJ}$\bs NP$_\text{SBJ}$}}\\
\bapply{6} & \permute{1}\\
\mc{6}{\text{NP$_\text{SBJ}$}} & \mc{1}{(\text{S}/\text{NP$_\text{OBJ}$})\bs\text{NP$_\text{SBJ}$}} \\
\bapply{7} & \\
\mc{7}{\text{S}/\text{NP$_\text{OBJ}$}} \\
\bapply{9}\\
\mc{9}{\text{S}}

}
\caption{Example with 9 words (\textsc{Medium} test)}
\end{subfigure}
\caption{Examples of sentences and their GCG derivation (somewhat simplified for space limitations).}
\label{fig:ccg_subexamples}
\end{figure*}

\section{Background}

\paragraph{Artificial Language Learning}
Artificial languages (ALs) are often used in the evaluation of LMs as they allow for more targeted evaluation of specific linguistic features.
Typical investigations include analyzing what kind of structure LMs can model, using ALs of varying complexities~\cite{suzgun2019lstm,weiss-et-al-2018-practical,el-naggar2022exploring,DBLP:conf/acl/KalliniPFMP24,DBLP:conf/coling/SomeyaYO24,DBLP:conf/iclr/DeletangRGGWCCH23}.
In addition to formal languages (e.g., Dyck languages), more linguistically grounded ALs, such as PCFG-based ones, have been developed to evaluate LM word-order inductive biases \cite{DBLP:conf/acl/WhiteC20}. 
\citet{el2025gcg,elnaggar2025wordordersfacilitatelength} extended PCFG-based ALs to generalized categorial grammar (GCG)-based ALs, which support a broader range of grammatical phenomena. 
ALs are also often used to test LM generalization under controlled conditions~\cite{weiss-et-al-2018-practical,suzgun2019lstm,el-naggar2022exploring,el2023theoretical,elnaggar2025wordordersfacilitatelength}.

\paragraph{Curriculum Learning}
Research has long questioned whether the order in which training data is exposed to neural network models affects their learning \cite{elman1991incremental,elman1993learning,rohde1999language,krueger2009flexible,bengio2009curriculum}.
\citet{elman1991incremental,elman1993learning} introduced the idea of ``starting small'' and conducted experiments where RNNs were trained in phases to learn English sentences, where sentence complexity was increased in each phase.
The starting small hypothesis has been revisited in several NLP applications, like learning and generalization of grammatical patterns \cite{rohde1999language}, unsupervised dependency parsing \cite{spitkovsky2009baby} and in a reinforcement learning framework \cite{krueger2009flexible}, and even in visual applications like shape recognition \cite{bengio2009curriculum}. 

\section{Experimental Settings}

\subsection{Original Setting}
\label{subsec:base_data}
We first briefly introduce the base datasets adapted from prior works~\cite{el2025gcg,elnaggar2025wordordersfacilitatelength}.
A set of GCG grammars is first defined via multiple independent word-order parameters generating 96 languages for the GCG-based AL corpus.
For example, one AL follows Japanese-like (head-final), and another follows English-like (mostly head-initial) word order.
Specifically, there are 7 binary parameters to determine (i) subject--verb order, (ii) object--verb order, (iii) subject--object order, (iv) complementizer--clause, (v) noun-adposition, (vi) noun-adjective, and (vii) position of relativizer. 
These configurations are denoted as a sequence of digits (e.g., \texttt{0001010}). 
Configurations with more \texttt{0}s tend to be head-final, while the ones with more \texttt{1}s are head-initial.
The binary parameter controls the directional slash of specific rules in the GCG grammar.
Examples of the sentence "Kim said that John touched Lisa" in grammars \texttt{0000000} (Japanese-like) and \texttt{0101101} (English-like) are:\footnote{Note that while the words in the ALs are pseudo-words, real English words are used in these examples for readability.}
\begin{itemize}[labelwidth=0pt,leftmargin=0pt,align=left,itemsep=0pt]
    \item[\texttt{0000000}:] \textit{Ken ga John ga Lisa o touch that said}
    \item[\texttt{0101101}:] \textit{Ken ga said that John ga touch Lisa o}
\end{itemize}
Note that the ALs have case markers for subject (\textit{ga}) and object (\textit{o}).
Figure~\ref{fig:ccg_subexamples} shows more examples of sentences derived by \texttt{0101101} (English-like) GCG grammar.
Note that sentences are first generated fully randomly, and then grammatical (parsable) ones are selected based on the GCG parser, where GCG parser configurations differ in different word order configurations.

In each AL, the training data consists of 80K sentences with 3--8 words (uniform length distribution), and the LMs' inductive biases are evaluated on three types of evaluation sets: (i) \textsc{Short} test set with the same length distribution as the training data, (ii) \textsc{Medium} test set with longer length distribution than the training set  (9-10) to test generalization, and (iii) \textsc{Long} test set that has further longer sentences created by several heuristics (11-20).
\footnote{These are available at \\\url{https://github.com/nadineelnaggar/gcg-based-artificial-languages}}
We follow this data split, but change how the training data is presented to the model and observe its effect on word-order preferences.
We define \textsc{Original} setting as the model training using this base dataset with a randomized data order.

\begin{figure*}[t]
    \centering
    \begin{subfigure}{1.0\linewidth}
        \centering
        \includegraphics[width=1.0\linewidth]{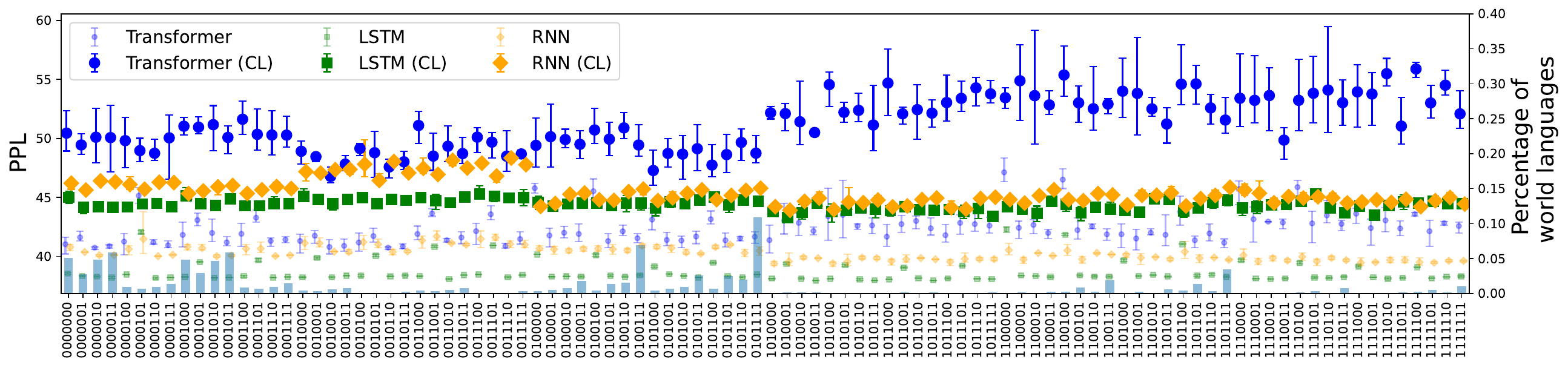}
        \caption{\textsc{Short} test (length 3--8).}
        \label{fig:perplexity-in-domain}
    \end{subfigure}

    \begin{subfigure}{1.0\linewidth}
        \centering
        \includegraphics[width=1.0\linewidth]{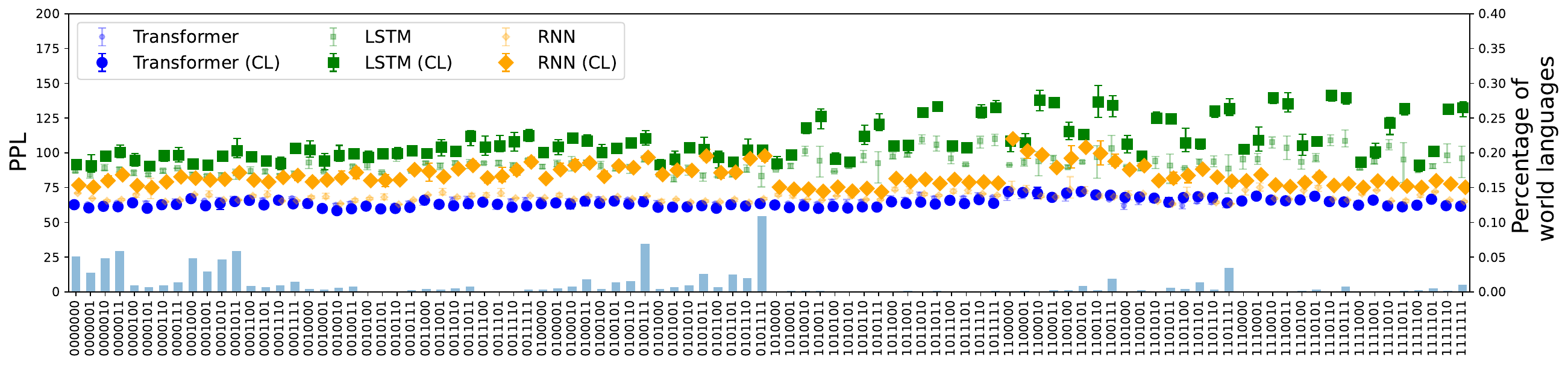}
        \caption{\textsc{Medium} test (length 9--10).}
        \label{fig:perplexity-out-domain}
    \end{subfigure}

    \begin{subfigure}{1.0\linewidth}
        \centering
        \includegraphics[width=1.0\linewidth]{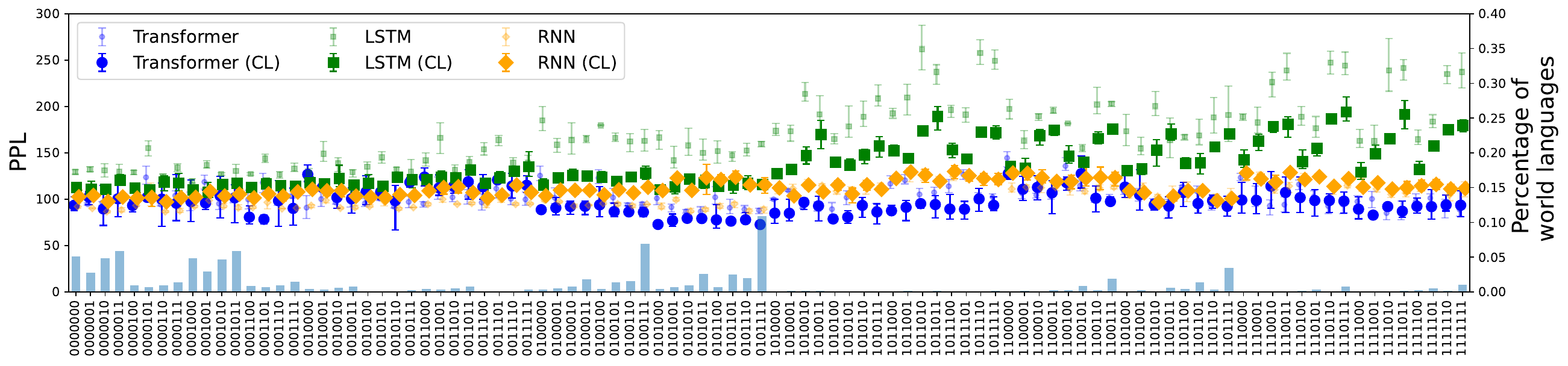}
        \caption{\textsc{Long} test (length 11--20).}
        \label{fig:perplexity-out-domain-long}
    \end{subfigure}
    
    \caption{Distributions of perplexities and typological plausibility across languages. The error bars indicate max and min PPLs within three runs. The smaller, semi-transparent markers correspond to the results without CL, which is excerpted from~\citet{elnaggar2025wordordersfacilitatelength}.}
    \label{fig:perplexity-comparison}
\end{figure*}

\subsection{Curriculum Learning Setting} 
\label{sec:dataset_creation}
To examine the CL effect on LMs' inductive biases, we introduce the CL setting in which the training data is presented based on a specific strategy.
As a case study, we explore the widely-used token length-based CL, and thus our questions will be: do LMs prefer specific word order configurations independently of the ordering of training data?
Or more generally, how reproducible are existing findings under different LM training scenarios?
We do not aim to find the best, optimal learning scenario to control LMs' learning preferences or to perform a comprehensive analysis, trying every possible learning strategy.

Specifically, we split the base training data into six parts (lengths of 3, 4, 5, 6, 7, and 8) with the same number of sentences in each part.
LMs are trained sequentially in stages corresponding to each training data part.
That is, the first training stage is with sentences of length 3, and the second stage will be with sentences of length 4, and so on.
To avoid catastrophic forgetting of previous data, we sample 5\% of the data from each past part and inject it into the current part.
For example, the data in the third stage consists of length-3 (5\%), length-4 (5\%), and length-5 (90\%) sentences.
For a fair comparison with the \textsc{Original} setting, the total number of training sentences is set to match the base dataset (80K).
For each of 96 ALs in the base dataset, we created a CL trained model.

\begin{table*}[t]
\centering
\begin{subtable}{\textwidth}
\centering
\begin{tabular*}{\textwidth}{@{\extracolsep{\fill}}llrrrrrrr@{}}
    \toprule
        & & \multicolumn{7}{c}{\textsc{Short}} \\
        \cmidrule(lr){3-9}
        Model & CL & SOV & OSV & SVO & OVS & VSO & VOS & TA $\downarrow$ \\
        \cmidrule(l){1-1} \cmidrule(l){2-2} \cmidrule(lr){3-8} \cmidrule(lr){9-9}
        
        Trans.~\cite{elnaggar2025wordordersfacilitatelength} 
        & & 41.8 & 41.6 & 42.3 & 42.6 & 42.7 & 43.3 
        & \textbf{$-$27.7}$^\dagger$ \\

        Trans. & \checkmark 
        & 50.2 & 48.8 & 49.3 & 52.6 & 53.3 & 53.4 
        & \textbf{$-$22.3}$^\dagger$ \\

        LSTM~\cite{elnaggar2025wordordersfacilitatelength} 
        & & 38.7 & 38.8 & 38.7 & 38.4 & 39.1 & 38.5 
        & \textbf{$-$14.2}$\:\:$ \\

        LSTM & \checkmark 
        & 44.4 & 44.9 & 44.5 & 43.9 & 44.2 & 44.4 
        & 16.1$\:\:$ \\

        RNN~\cite{elnaggar2025wordordersfacilitatelength}  
        & & 40.4 & 41.0 & 40.6 & 39.7 & 40.1 & 39.7 
        & 13.0$\:\:$ \\

        RNN  & \checkmark 
        & 45.9 & 47.4 & 45.1 & 44.5 & 45.0 & 44.8 
        & 14.2$\:\:$ \\

        \cmidrule(lr){1-9}
        NL (Prob. $\uparrow$) 
        & & 0.54 & 0.04 & 0.23 & 0.01 & 0.12 & 0.05 & - \\
        
        \bottomrule
    \end{tabular*}
    \caption{\textsc{Short} test set.}
\end{subtable}

\begin{subtable}{\textwidth}
\centering
\begin{tabular*}{\textwidth}{@{\extracolsep{\fill}}llrrrrrrr@{}}
\toprule
& & \multicolumn{7}{c}{\textsc{Medium}} \\
\cmidrule(lr){3-9}
Model & CL & SOV & OSV & SVO & OVS & VSO & VOS & TA $\downarrow$ \\
\cmidrule(l){1-1} \cmidrule(l){2-2}
\cmidrule(lr){3-8} \cmidrule(lr){9-9}

Trans.~\cite{elnaggar2025wordordersfacilitatelength}
& & 65.2 & 63.5 & 64.2 & 65.9 & 66.1 & 65.0
& \textbf{$-$10.4}$\:\:$ \\

Trans. & \checkmark
& 63.2 & 61.8 & 62.9 & 62.8 & 68.9 & 64.7
& \textbf{$-$5.9}$\:\:$ \\

LSTM~\cite{elnaggar2025wordordersfacilitatelength}
& & 85.9 & 91.7 & 88.0 & 97.5 & 92.9 & 97.9
& \textbf{$-$31.0}$^\dagger$ \\

LSTM & \checkmark
& 95.6 & 102.3 & 101.9 & 112.4 & 119.8 & 117.7
& \textbf{$-$20.1}$^\dagger$ \\

RNN~\cite{elnaggar2025wordordersfacilitatelength}
& & 67.8 & 67.9 & 66.7 & 69.6 & 69.0 & 69.4
& \textbf{$-$17.4}$\:\:$ \\

RNN & \checkmark
& 80.3 & 84.5 & 89.7 & 76.6 & 91.7 & 78.1
& 21.2$\:\:$ \\

\cmidrule(lr){1-9}
NL (Prob. $\uparrow$)
& & 0.54 & 0.04 & 0.23 & 0.01 & 0.12 & 0.05 & - \\
\bottomrule
\end{tabular*}
\caption{\textsc{Medium} test set.}
\end{subtable}

\begin{subtable}{\textwidth}
\centering
\begin{tabular*}{\textwidth}{@{\extracolsep{\fill}}llrrrrrrr@{}}
\toprule
& & \multicolumn{7}{c}{\textsc{Long}} \\
\cmidrule(lr){3-9}
Model & CL & SOV & OSV & SVO & OVS & VSO & VOS & TA $\downarrow$ \\
\cmidrule(l){1-1} \cmidrule(l){2-2}
\cmidrule(lr){3-8} \cmidrule(lr){9-9}

Trans.~\cite{elnaggar2025wordordersfacilitatelength}
& & 102.3 & 99.4 & 97.9 & 104.0 & 107.6 & 97.9
& \textbf{$-$19.2}$\:\:$ \\

Trans. & \checkmark
& 95.1 & 112.9 & 83.1 & 89.9 & 106.2 & 96.2
& \textbf{$-$18.0}$^\dagger$ \\

LSTM~\cite{elnaggar2025wordordersfacilitatelength}
& & 131.9 & 141.5 & 160.7 & 205.5 & 180.9 & 207.5
& \textbf{$-$33.4}$^\dagger$ \\

LSTM & \checkmark
& 113.8 & 122.0 & 119.3 & 153.6 & 151.9 & 163.7
& \textbf{$-$32.3}$^\dagger$ \\

RNN~\cite{elnaggar2025wordordersfacilitatelength}
& & 91.8 & 94.6 & 93.2 & 118.0 & 109.0 & 114.2
& \textbf{$-$43.1}$^\dagger$ \\

RNN & \checkmark
& 102.9 & 107.2 & 113.0 & 117.6 & 113.7 & 117.8
& \textbf{$-$20.2}$^\dagger$ \\

\cmidrule(lr){1-9}
NL (Prob. $\uparrow$)
& & 0.54 & 0.04 & 0.23 & 0.01 & 0.12 & 0.05 & - \\
\bottomrule
\end{tabular*}
\caption{\textsc{Long} test set.}
\end{subtable}

\caption{Average PPLs within each base word order group and Pearson's correlation coefficient (TA) between PPL and typological frequency. Negative TA scores are highlighted in bold. Statistical significance (p<0.05) is marked with $\dagger$. NL is the percentage of natural languages possessing each word order.}
\label{tab:generalization_medium_long_split}
\label{tab:main_tables}
\end{table*}

\begin{figure*}[t]
  \centering

  \begin{subfigure}{0.38\linewidth}
    \centering
    \includegraphics[width=\linewidth]{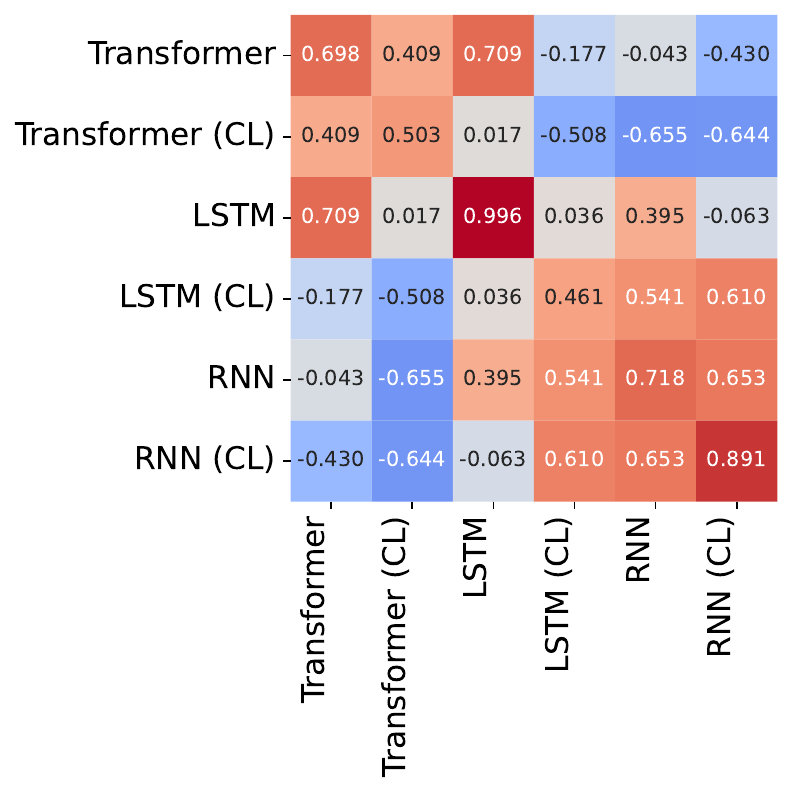}
    \caption{\textsc{Short}}
    \label{fig:sub1}
  \end{subfigure}
 \hspace{0pt}
  \begin{subfigure}{0.25\linewidth}
    \centering
    \includegraphics[width=\linewidth]{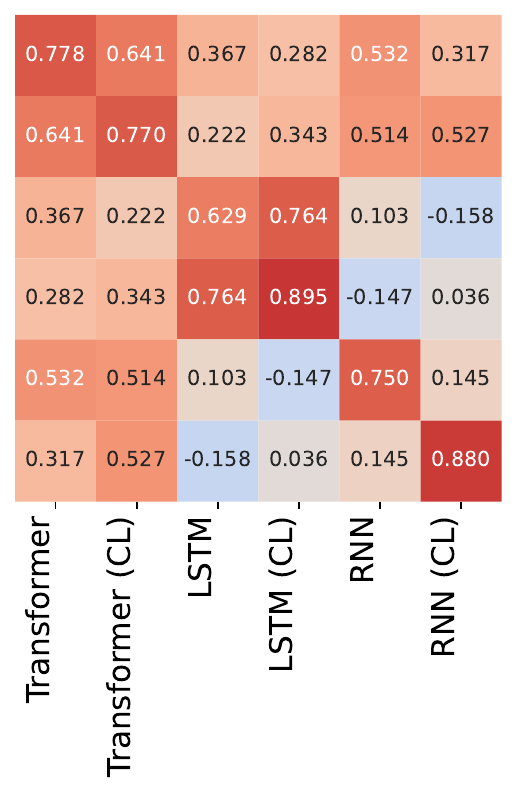}
    \caption{\textsc{medium}}
    \label{fig:sub2}
  \end{subfigure}
   \hspace{0pt}
  \begin{subfigure}{0.33\linewidth}
    \centering
    \includegraphics[width=\linewidth]{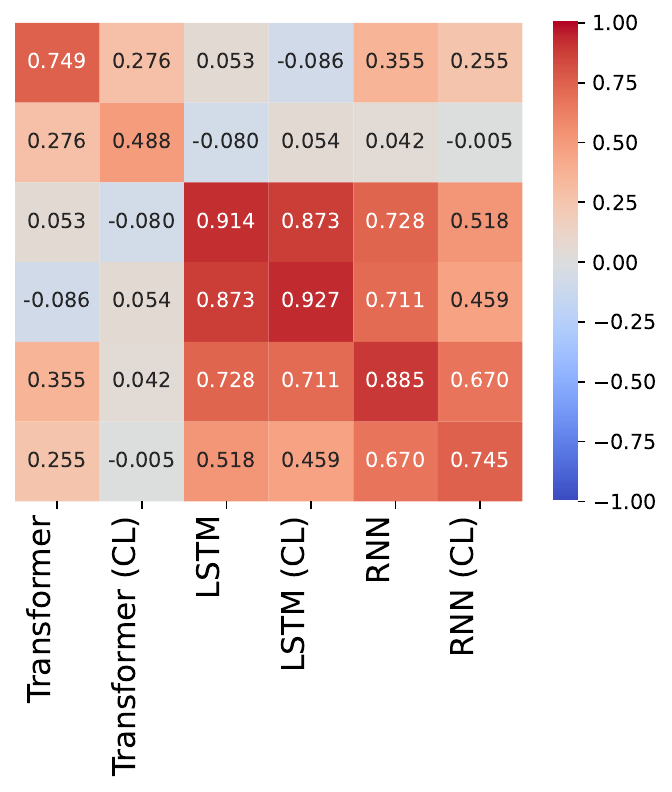}
    \caption{\textsc{Long}}
    \label{fig:sub3}
  \end{subfigure}

  \caption{Correlation of word order preference between different models$\times$CL. Transformer (CL), for example, denotes Transformer-based LM trained with CL. Orthogonal elements denote seed variance, i.e., average correlations between the same model but different seeds.}

  \label{fig:correlations}
\end{figure*}

\section{Experiments}
We adopt three evaluation settings and replicate~\citet{elnaggar2025wordordersfacilitatelength}.
For non-CL, \textsc{Original} settings, results are excerpted from that study for a comparison, and our new CL-based results are superimposed on them.
We focus on RNN, LSTM, and Transformer LMs.

\paragraph{Model Training.}
For the CL settings, we adopt the same setting as the original ones, except for CL (see Appendix~\ref{appendix:model_hyperparameters} for hyperparameter details).
We train LMs with three different random seeds for each model, using the Fairseq toolkit \cite{ott2019fairseq}.
Under CL, LMs are trained for 2 epochs in each stage, except the last, and the optimizer (e.g., learning rate scheduler) is inherited across all the stages. 
At the last stage, an early-stopping criterion of 5-epoch patience is adopted (i.e., if the validation PPL is not improved for 5 epochs, the training stops), following the non-CL settings.

\paragraph{Evaluation.}
The \textsc{Short} test set evaluates the LMs' in-distribution modeling ability, and  \textsc{Medium} and \textsc{Long} ones evaluate the out-of-distribution generalization ability of LMs.
Note that the sentences in the \textsc{Short} test sets are unseen in the training data.
We report average perplexity (PPL) among three runs for each word order configuration.
We also report typological alignment (TA) scores --- Pearson's correlation coefficient between model's average PPL and the frequency of respective word order configurations in the world~\cite{DBLP:conf/acl/KuribayashiUYOB24,el2025gcg,elnaggar2025wordordersfacilitatelength}.
A negative TA score reflects a better alignment of a LM's learning preference with typological tendencies, i.e., typologically common word orders are easier to learn for the LMs.
Our focus is on how the language-learning curriculum (\textsc{Original} vs. CL) affects their learning preferences (PPLs and TAs).

\subsection{Experiment 1: PPLs}
\label{ssec:exp1}
\paragraph{PPL variations and CL.}
We first evaluate LMs based on perplexity (PPL) on the \textsc{Short}, \textsc{Medium}, and \textsc{Long} test sets.
Figure~\ref{fig:perplexity-comparison} shows the general PPL tendencies (y-axis) over different word order configurations (x-axis) in each test set.
Here, PPLs from LMs without CL are also shown with smaller, semi-transparent markers.\footnote{These scores without CL are exactly the same as those reported in~\citet{elnaggar2025wordordersfacilitatelength}. Our CL results are comparable with their study as the adaptation of CL is the only difference.}
As a quick check, we first confirm that word order preference still varies even when data are presented in a comprehensive way with CL.
Compared to the original results (smaller, semi-transparent markers in Figure~\ref{fig:perplexity-comparison}), we observe that LMs with CL tend to exhibit worse PPL on \textsc{Short} tests, but comparable or better PPL on \textsc{Medium} and \textsc{Long} tests, suggesting less overfitting to short sentences and a positive effect of CL for length generalization.

\paragraph{Typological variations and CL.}
Table~\ref{tab:main_tables} aggregates the results of our CL-based LMs, and the TA score is also computed.
To summarize the results, first, the TA scores, especially in  RNN/LSTM results on \textsc{Medium}/\textsc{Short} sets, substantially change with and without CL, and this basically leads to a worse TA.
Second, the TA scores in \textsc{Long} set are still significantly negative; that is, typologically common word order configurations facilitate generalization to longer sentences, regardless of CL.
Therefore the learning setting does affect the results, and in our case, this impact was not equally strong nor random across models/conditions, but rather altered specific results on individual ALs and test sets. Further experimentation with different approaches to CL, such as ordering by construction complexity as opposed to token length, may shed further light on the interactions between models, learning scenarios and inductive bias.

\paragraph{Inter-model correlations.}
Figure~\ref{fig:correlations} shows inter-correlation of PPL-vectors over 96 word orders from different LMs and CL settings.
Note that the orthogonal elements in the matrices show the average correlation between PPL-vectors from the same LM architecture, but with different seeds.
First, PPL correlation between the same model with and without CL is generally smaller than the seed variance (respective orthogonal elements), showing that CL does affect the word order preference of LMs.
Second, CL has different effects depending on LM architectures; for example, in \textsc{Short} test, LSTM vs. LSTM (CL) shows nearly zero correlation (0.036), in \textsc{Medium} test, RNN vs. RNN (CL) shows a small correlation (0.145), while in \textsc{Long} test, Transformer vs. Transformer (CL) shows a relatively small correlation (0.276).
This demonstrates subtle and non-obvious interactions between word order preferences, model architectures, and learning scenarios.

\begin{table*}[t]
    \centering
    \footnotesize
    \tabcolsep 0.07cm
\begin{tabular*}{\textwidth}{@{\extracolsep{\fill}}llrrrrrr@{}}
    \toprule
    & & \textsc{Recursive} & \textsc{Embedded}  
      & \multicolumn{2}{c}{Case Judgment}
      & \multicolumn{2}{c}{Verb Judgment} \\
    \cmidrule(lr){3-4} \cmidrule(lr){5-6} \cmidrule(lr){7-8}
    Model & CL 
    & TA ($\downarrow$)
    & TA ($\downarrow$)
    & Acc. ($\uparrow$)
    & Corr. ($\uparrow$)
    & Acc. ($\uparrow$)
    & Corr. ($\uparrow$)
    \\
    \midrule
    
    Trans.~\cite{elnaggar2025wordordersfacilitatelength}
        &  & $-$5.1\phantom{$^\dagger$} & $-$23.5$^\dagger$
        & 97.7$\pm$1.5\phantom{$^\dagger$} & 0.14$\:$\phantom{$^\dagger$}
        & 81.0$\pm$14.7\phantom{$^\dagger$} & 0.27$^\dagger$ \\
    Trans.
        & \checkmark
        & 5.5\phantom{$^\dagger$} & $-$19.2$^\dagger$
        & 96.3$\pm$3.1\phantom{$^\dagger$} & 0.01$\:$\phantom{$^\dagger$}
        & 74.4$\pm$22.2\phantom{$^\dagger$} & 0.26$^\dagger$ \\
    \cmidrule(lr){1-8}
    
    LSTM~\cite{elnaggar2025wordordersfacilitatelength}
        &  & 9.2\phantom{$^\dagger$} & $-$3.7\phantom{$^\dagger$}
        & 97.2$\pm$1.4\phantom{$^\dagger$} & 0.03$\:$\phantom{$^\dagger$}
        & 85.1$\pm$9.6\phantom{$^\dagger$} & 0.28$^\dagger$ \\
    LSTM
        & \checkmark
        & 16.2\phantom{$^\dagger$} & 0.4\phantom{$^\dagger$}
        & 95.6$\pm$1.9\phantom{$^\dagger$} & -0.31$^\dagger$
        & 78.1$\pm$11.7\phantom{$^\dagger$} & 0.32$^\dagger$ \\
    \cmidrule(lr){1-8}
    
    RNN~\cite{elnaggar2025wordordersfacilitatelength}
        &  & 12.9\phantom{$^\dagger$} & $-$18.1$^\dagger$
        & 97.4$\pm$1.4\phantom{$^\dagger$} & 0.21$^\dagger$
        & 77.4$\pm$15.5\phantom{$^\dagger$} & 0.23$^\dagger$ \\
    RNN
        & \checkmark
        & 17.0\phantom{$^\dagger$} & $-$18.9$^\dagger$
        & 92.6$\pm$5.8\phantom{$^\dagger$} & 0.10$^\dagger$
        & 63.0$\pm$25.1\phantom{$^\dagger$} & 0.31$^\dagger$ \\
    
    \bottomrule
    \end{tabular*}
    \caption{Correlation between PPLs and word-order frequencies in targeted generalization sets (Recursive/Embedded) and correlation between accuracy and word-order frequencies in grammaticality judgment tests. Statistical significance (p<0.05) is marked with $\dagger$.}
    \label{tab:analysis}
\end{table*}

\subsection{Experiment 2: Targeted Evaluations}
\label{ssec:exp2}
We also replicate the targeted evaluations conducted in~\citet{elnaggar2025wordordersfacilitatelength} (see details in their Sections 6 and 7). 

\subsubsection{PPLs in Targeted Generalization Sets.}
We measure PPLs on the test data with specific complex constructions.
Here, the generalization is evaluated on unbounded dependency structures, specifically recursive relative clauses (\textsc{Recursive}), where relative clauses are nested, and embedded relative clauses (\textsc{Embedded}), where a relative clause is embedded in a subordinate clause. 
English examples are as follows:
\begin{description}
    \item[Recursive Relative Clauses:] \textit{fruits ga which pasta ga which John ga promised nibbles received wall o}
    \item[Embedded Relative Clause:] \textit{fruits ga which John ga said that pasta ga nibbles received wall o}
\end{description}

In the \textsc{Recursive} test set, two relative clauses are nested.
In the \textsc{Embedded} test set, a relative clause is embedded in a subordinate clause, e.g., ``he said''. 
Through randomly sampling lexical items for the above two fixed construction templates, 500 examples for each construction are generated in each of 96 word order variations.
Note that the resulting sentences are longer than 8 tokens (training data).
PPLs are measured in each targeted test set, and these are aggregated as the TA score, correlation between average PPL over three model with different random seeds and word order frequency across 96 languages. 

Table~\ref{tab:analysis} (left-side) shows 
results with and without CL. 
CL typically yields comparable or slightly worse TA scores, while the relative order of TA scores among conditions is basically preserved, e.g., Transformer in \textsc{Embedded} yields the lowest TA, while RNN in \textsc{Recursive} yields the highest TA with or without CL.

\subsubsection{Grammaticality Judgments.}
We also perform grammaticality judgment evaluation with two test suites: (i) case-type accuracy, and (ii) verb-type accuracy. 
English-like (\texttt{0101101}) examples are as follows:
\begin{description}[itemsep=0pt]
\item[Case-type judgment:]\footnote{\textit{ga} and \textit{o} in the examples are subject and object case markers, respectively.}$\;$
\begin{itemize}[itemsep=0pt]
    \item \textit{fluffy soft and intelligent mango \ul{ga} controls
owl o}
\item *\textit{fluffy soft and intelligent mango \ul{o} controls
owl o}
\end{itemize}
\item[Verb-type judgment:]\footnote{\textit{escorts} and \textit{evolves} in the examples are transitive and intransitive verbs, respectively.} $\;$
\begin{itemize}[itemsep=0pt]
    \item \textit{scooter ga which green machine ga \ul{escorts} 
walk}
\item *\textit{scooter ga which green machine ga \ul{evolves}
walk}
\end{itemize}
\end{description}

For the case-type judgment, a randomly selected grammatical case marker is replaced with an ungrammatical one, e.g. ``ga'' is replaced with ``o'' or vice versa. 
For verb-type judgment, a transitive verb in a grammatical sentence is replaced with an intransitive verb, resulting in the sentence becoming ungrammatical (marked with * in the examples above).
For each word order, 500 items are sampled from the \textsc{Medium} test set, and their ungrammatical variants are created by applying specific modifications.
Should there be more than one potential replacement candidate in a sentence, one is chosen at random. 
We report accuracy (Acc.) of grammatical judgments and correlation (Corr.) with typological commonality --- the higher the better.

Table~\ref{tab:analysis} (right side) shows the results. 
CL consistently worsened the accuracies, and the typological alignments are also weakened (Case-type judgment) or comparable (Verb-type judgment). 
The non-positive effect of CL is similar to the other experiments.

\section{Discussion}
\label{sec:discussion}
Overall, CL led to a slightly negative effect for typological alignment between learnability and typological patterns, which leads to some interpretations.
First, the results empirically demonstrated the considerable interaction effects between LMs' word-order preferences, or more generally, inductive biases, and how data are presented.
It is worth revisiting LMs' learning biases in developmentally motivated language-learning scenarios.
Second, if the length-based CL we adopted is truly a good proxy for human language acquisition, the alignment between LMs' inductive biases and typological patterns might have been overestimated in existing non-CL studies.
This view, more or less, challenges the idea that learners' biases shape language design and, indirectly, evokes the dominance of alternative hypotheses associating language design with cognitive-external factors, e.g., historical accidents or geographical influence~\cite{moravcsik1978language,Atkinson2011-kt}; see~\citet{Culbertson2012-bo} i.a. for a more comprehensive overview.
Third, otherwise, if we stand on the premise of the learner's bias shaping language design, the weaker typological alignment under the adopted length-based CL raises the concerns that the  length-based control is too simplified CL strategy, given broader possibility of CL implementations~\cite{Wang2022-fm}, or CL alone is insufficient to simulate the plausible biases in language acquisition~\cite{Perfors2012-km}. 
These views at least poses the next question: are there any other CL strategies or factors in addition to CL leading to better simulating the emergence of typological patterns?

\section{Conclusion}
\label{sec:conclusion}
In this paper, we revisit the prior work~\cite{elnaggar2025wordordersfacilitatelength} by introducing one basic type of curriculum learning (CL).
In our experiments, we ablate the CL effect on the inductive bias of RNN, LSTM, and Transformer LMs towards different word-order configurations.
We confirm that the LMs' word-order preferences indeed change under CL, suggesting that the order of the training data affects apparent LM learning biases.
Our results show subtle and somewhat inconsistent interactions between individual ALs, model architectures, learning scenarios, and inductive bias. 
In the future, we can explore several directions to extend this line of work and gain further insights. 
One of them is to investigate different CL strategies, such as the construction-complexity approach or model-based control of working memory growth~\cite{Wang2022-fm,mita-etal-2025-developmentally}.

\section*{Limitations}
In addition to the future directions mentioned in~\cref{sec:conclusion}, the definition of typological alignment should be carefully addressed.
For example, we have used Pearson's correlation to examine the relationship between typological commonality and perplexity in this study, following~\citet{DBLP:conf/acl/KuribayashiUYOB24} and \citet{elnaggar2025wordordersfacilitatelength}, but its validity should be re-investigated.
Finally, this study explanatorily analyzed the effect of CL on LMs' inductive bias evaluation and does not derive any strong conclusion like CL clearly promotes/hinders typological alignment between LMs' learning biases and typological commonalities.
We rather hope that our paper introduces and encourages this interdisciplinary exploration space combining LMs' training scenarios and typological alignment, particularly given the progressive interdisciplinary integration between language modeling and language science.

\section*{Ethics Statement}
The data used in this paper is all artificial data based mostly on English words. 
We have no ethical concerns with the contents of this paper.

\section*{Acknowledgment}
We would like to thank the anonymous reviewers for their insightful feedback.

\section*{References}
\bibliographystyle{lrec2026-natbib}
\bibliography{lrec2026-example-bib}

\appendix
\section{Categorial Grammar}
\begin{figure*}
   
    \begin{subfigure}[t]{\textwidth}
    \small
    \centering
        \colorbox{SkyBlue}{funny Kim and happy Sandy ga met Felix o}\colorbox{orange}{and}\colorbox{SpringGreen}{Tom and Jerry ga caused trouble o}
        \caption{Example of two valid sentences are concatenated with a conjunction to create a longer one.}
        \label{fig:examples_template_extension_append_and}
    \end{subfigure}
    \\
    \begin{subfigure}[t]{\textwidth}
    \small
    \centering
    \colorbox{SkyBlue}{funny Kim}\colorbox{orange}{and}\colorbox{SpringGreen}{Tom and Jerry ga caused trouble o}\colorbox{SkyBlue}{and happy Sandy ga met Felix o}
    \caption{Example of how a valid sentence is embedded into another valid sentence to create a longer one.
    }
        \label{fig:examples_template_extension_embed}
    \end{subfigure}
    \caption{Examples of short templates (lengths 3-10) being combined to create longer templates (lengths 11-20).
    The first sentence is in blue, the second sentence is in green, and the conjunction is in orange. 
    }

    \label{fig:examples_template_extension}
\end{figure*}

\begin{figure*}
\centering
\begin{subfigure}[t]{0.3\textwidth}

    \includegraphics[width=\textwidth]{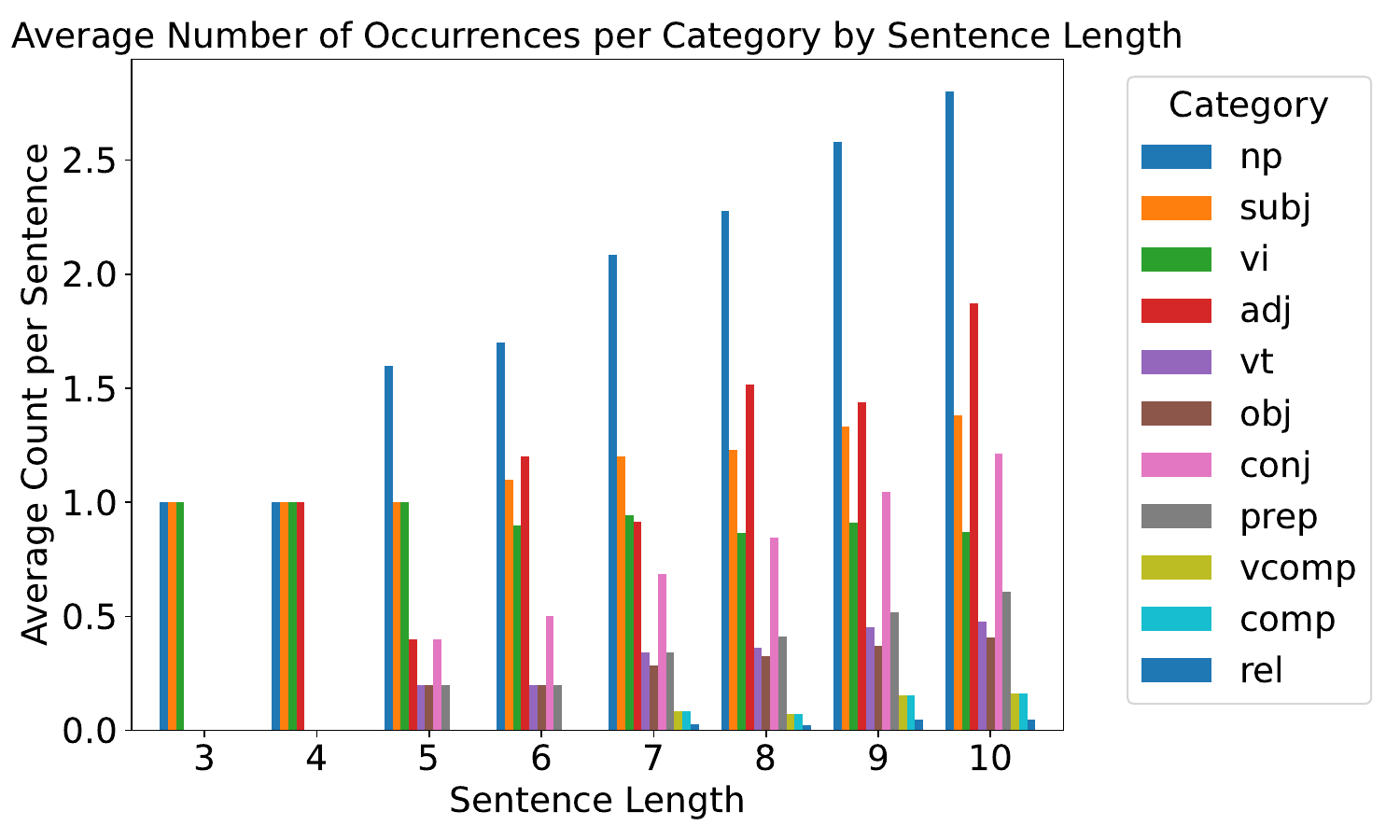}
    \caption{How many times each category appears in \textsc{Short} and \textsc{Medium} per template on average.}
    \label{fig:template-avg-category-count-per-length}
\end{subfigure}
\hfill
\begin{subfigure}[t]{0.3\textwidth}

    \includegraphics[width=\textwidth]{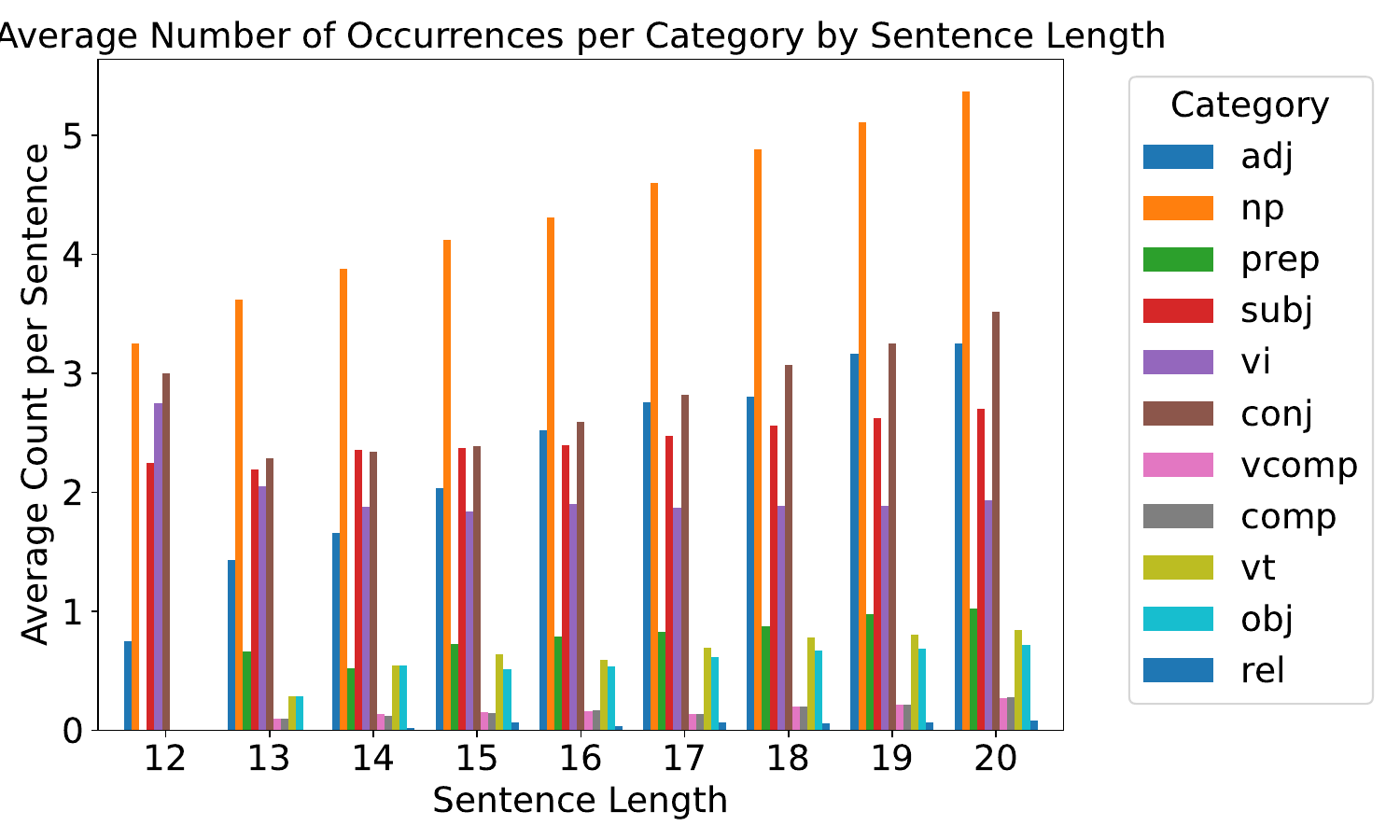}
    \caption{How many times each category appears in templates used to create \textsc{Long} set, per template on average.}
    \label{fig:template-avg-category-count-per-length-long-sampled}
\end{subfigure}
\hfill
\begin{subfigure}[t]{0.3\textwidth}

    \includegraphics[width=\textwidth]{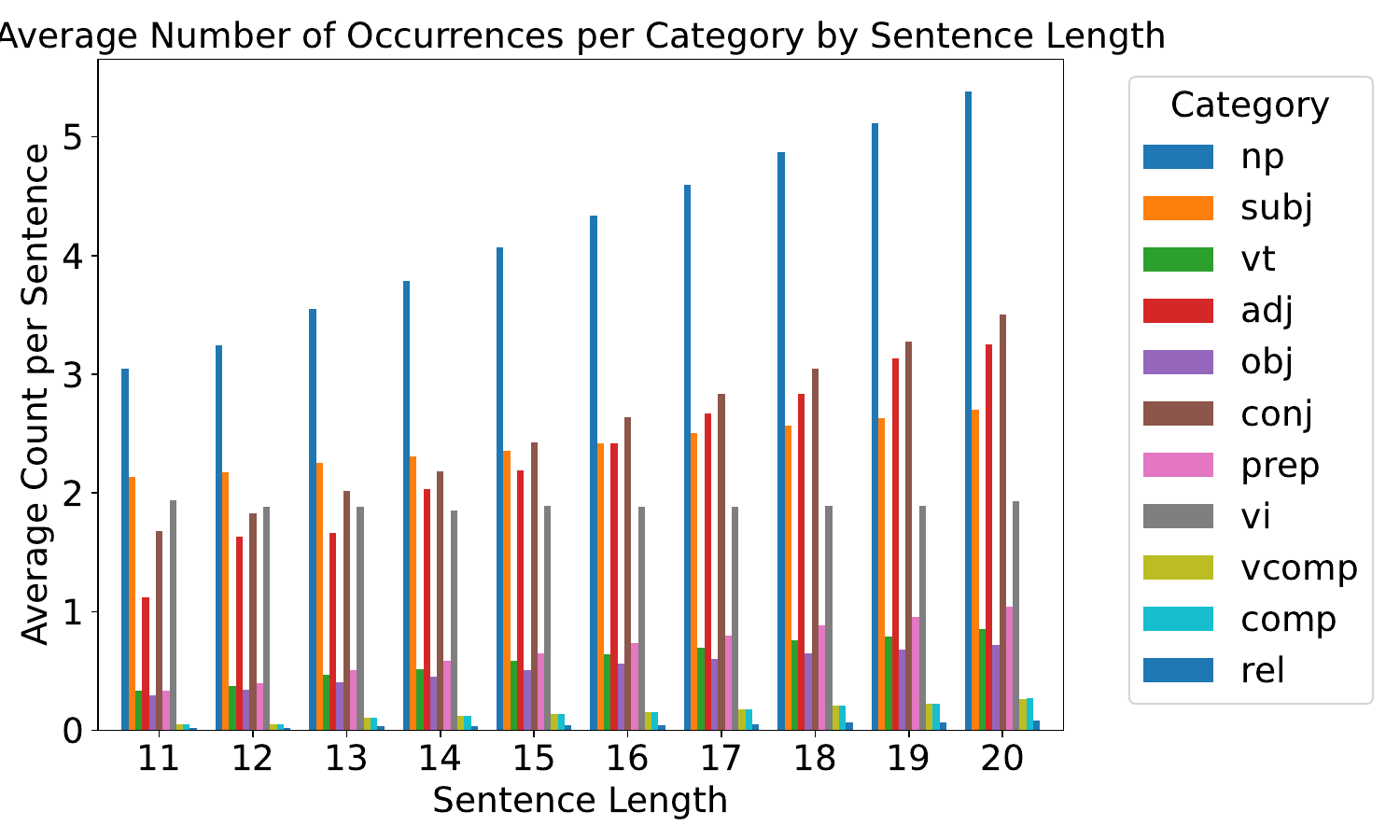}
    \caption{How many times each category appears in \textsc{Long} per template on average.}
    \label{fig:template-avg-category-count-per-length-long-all}
\end{subfigure}
\caption{The number of occurrences of the different categories for all templates lengths 3-10 (a), the templates used to create the long dataset (b), and all the generated long templates (c).}
\label{fig:avg-category-count-per-length-short-long}
\end{figure*}

\begin{figure*}[ht!]
    \begin{subfigure}[t]{0.3\textwidth}
    \includegraphics[width=\textwidth]{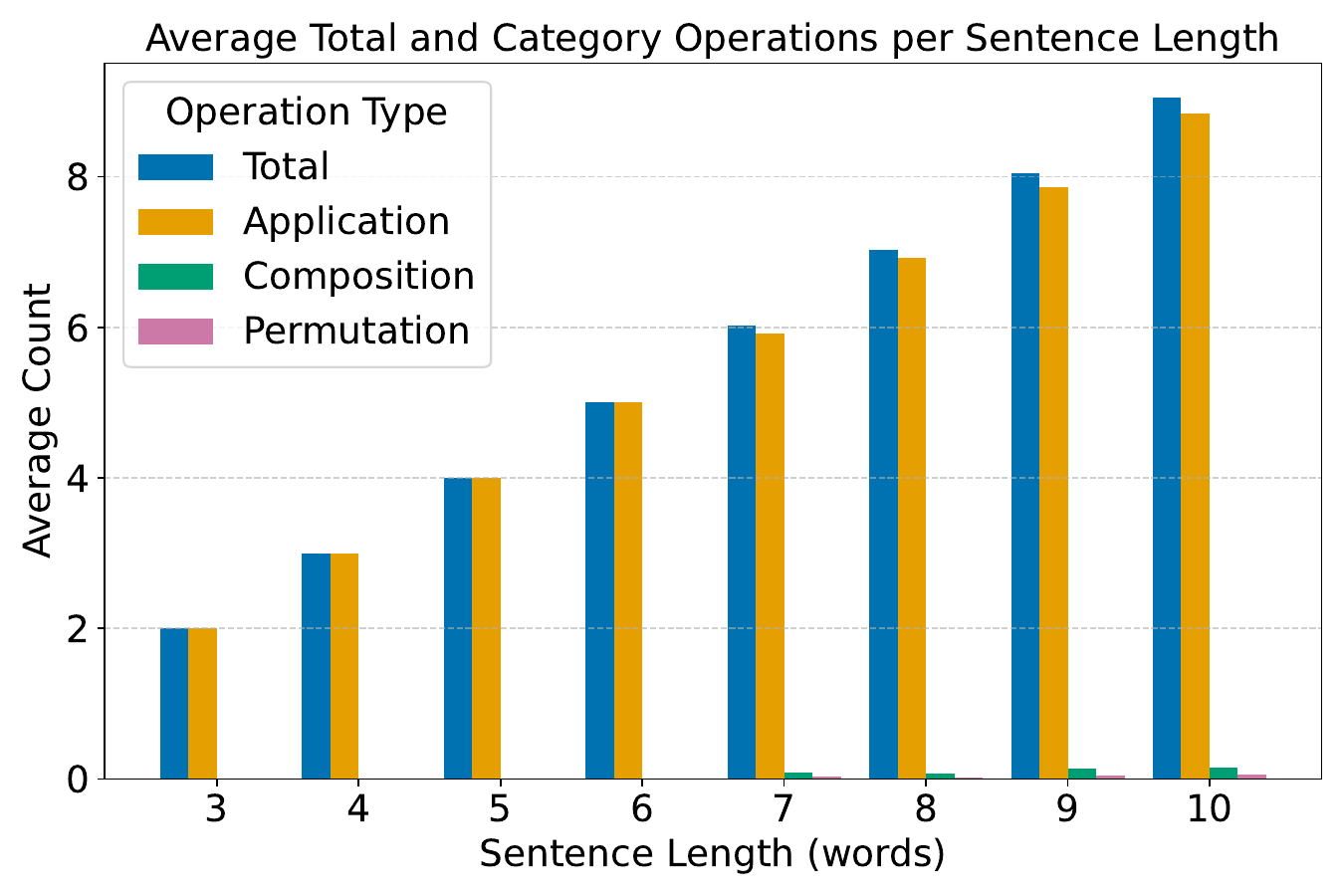}
        \caption{Template lengths 3-10 words (\textsc{Short} and \textsc{Medium}).}
        \label{fig:operations_3to10}
    \end{subfigure}
    \hfill
    \begin{subfigure}[t]{0.3\textwidth}
    \includegraphics[width=\textwidth]{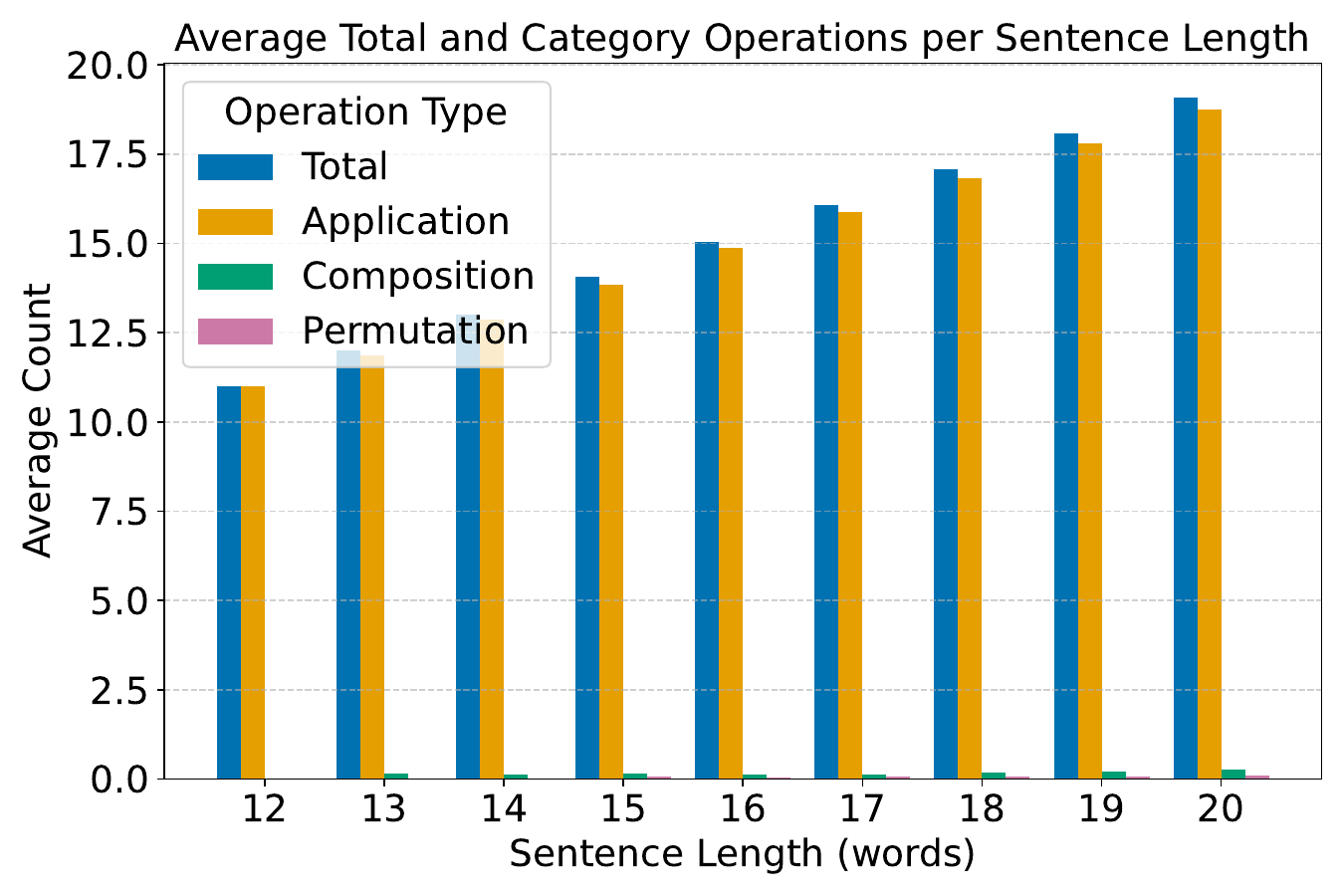}
    \caption{Sampled templates used to create the \textsc{Long} dataset.}
    \label{fig:operations_11to20_sampled}
    \end{subfigure}
    \hfill
    \begin{subfigure}[t]{0.3\textwidth}
    \includegraphics[width=\textwidth]{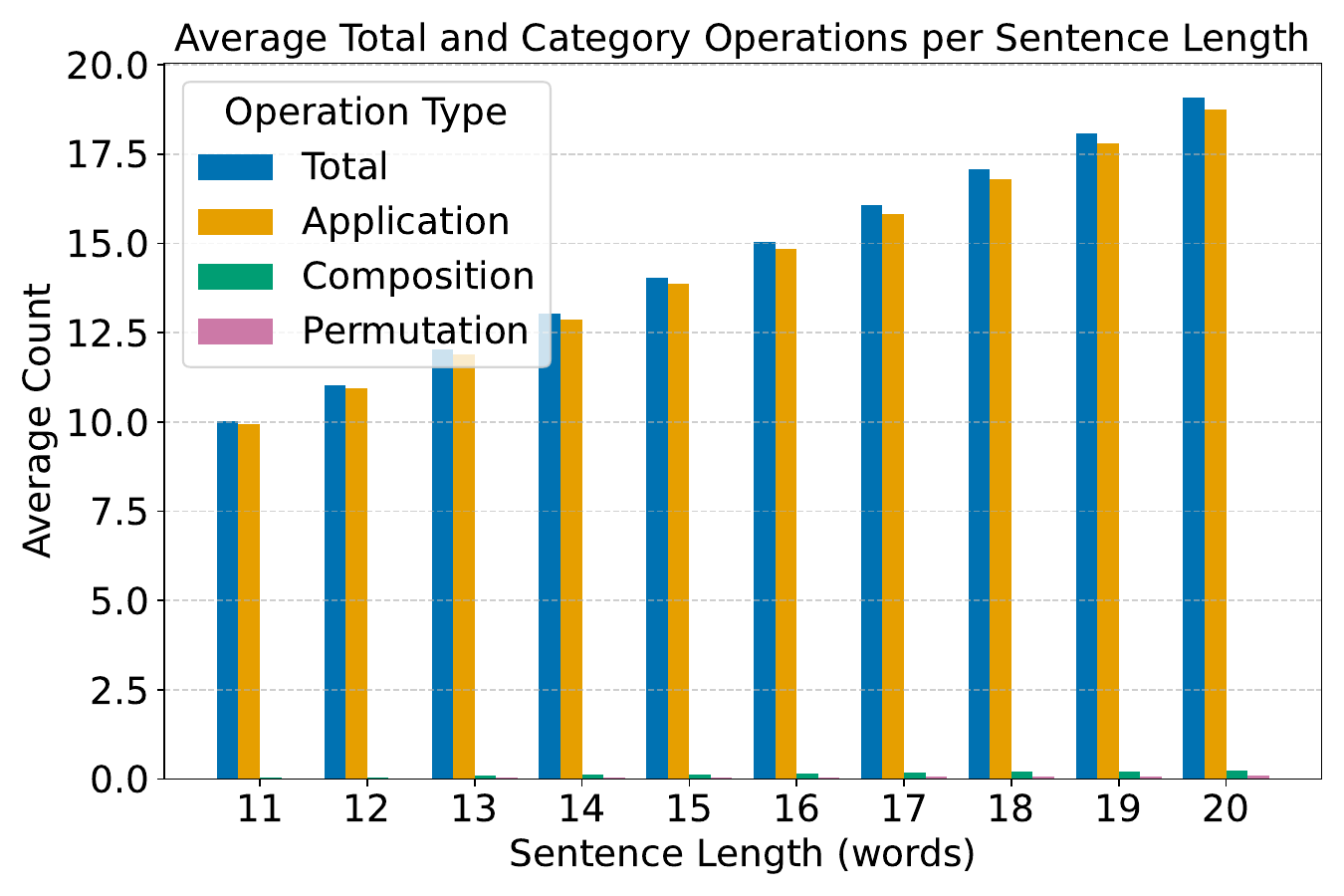}
    \caption{All extended templates created of lengths 11-20.}
    \label{fig:operations_11to20_all}
    \end{subfigure}
    \caption{The average number of combinatory operations in the GCG derivation for the templates of different lengths for the language configuration \texttt{0101101} by \citet{elnaggar2025wordordersfacilitatelength}. }

    \label{fig:operations_count}
\end{figure*}

Categorial Grammar (CG) is a formalism consisting of a lexicon, where each word is assigned a functor category, and a set of rules that define how the categories combine with each other. 
CG was introduced with the combinatory operation \textbf{application}, which has forward and backward variants.
Extensions of CG have been introduced, such as combinatory categorial grammar (CCG) \cite{steedman1996surface} and generalized categorial grammar (GCG) \cite{wood2014categorial}.
\citet{elnaggar2025wordordersfacilitatelength} use a GCG with the following combinatory operations, which they detail in their paper: 
\begin{itemize}[itemsep=0pt]
    \item Application (forward and backward variants)
    \item Coordination
    \item Composition (forward and backward variants)
    \item Weak Generalized Permutation 
\end{itemize}
This GCG allows for the expression of mildly context-sensitive constructions, like cross-serial and unbounded dependencies \cite{briscoe1997co,briscoe2000grammatical}.
Furthermore, by using GCG-based ALs in experiments, we are able to test a wider attested range of word order configurations including VSO and OSV base word order ALs.

\section{Dataset Details}
\label{appendix:dataset_details}
\begin{table*}[ht]
    \centering
    \small
\begin{minipage}[t]{\hsize}
\renewcommand{\arraystretch}{0.4}
    \centering
    \begin{tabular}{p{3cm}p{5cm}p{4.5cm}} \toprule
     \multirow{10}{1cm}{Fairseq model}
      & share-decoder-input-output-embed & True \\
      & embed\_dim & 128 \\
      & ffn\_embed\_dim & 512 \\
      & layers & 2 \\
      & heads & 2 \\
      & dropout & 0.3 \\
      & attention\_dropout & 0.1 \\
      & \#params. & 462K \\
    \cmidrule(lr){1-1} \cmidrule(lr){2-2} \cmidrule(lr){3-3}
    \multirow{5}{*}{Optimizer} & algorithm & AdamW \\
    & learning rates & 5e-4 \\
    & betas & (0.9, 0.98) \\
    & weight decay & 0.01 \\
    & clip norm & 0.0 \\
    \cmidrule(lr){1-1} \cmidrule(lr){2-2} \cmidrule(lr){3-3}
    \multirow{3}{3cm}{Learning rate scheduler} & type & inverse\_sqrt \\
    & warmup updates & 400 \\
    & warmup init learning rate & 1e-7 \\
    \cmidrule(lr){1-1} \cmidrule(lr){2-2} \cmidrule(lr){3-3}
    \multirow{4}{*}{Training}   
    & batch size & 2,048 tokens \\
    & tokens-per-sample & 128 tokens \\
    & sample-break-mode & none \\
    & epochs & 10 \\ 
    \bottomrule
        \end{tabular}
        \subcaption{Transformer.}
        \label{tbl:hyperparam_tl}
\end{minipage}

\begin{minipage}[t]{\hsize}
\renewcommand{\arraystretch}{0.4}
    \centering
    \begin{tabular}{p{3cm}p{5cm}p{4.5cm}} \toprule
     \multirow{8}{1cm}{Fairseq model} 
      & share-decoder-input-output-embed & True \\
      & embed\_dim & 128 \\
      & hiden\_size & 512 \\
      & layers & 2 \\
      & dropout & 0.1 \\
      & \#params. & 3,547K \\
    \cmidrule(lr){1-1} \cmidrule(lr){2-2} \cmidrule(lr){3-3}
    \multirow{5}{*}{Optimizer} & algorithm & AdamW \\
    & learning rates & 5e-4 \\
    & betas & (0.9, 0.98) \\
    & weight decay & 0.01 \\
    & clip norm & 0.0 \\
    \cmidrule(lr){1-1} \cmidrule(lr){2-2} \cmidrule(lr){3-3}
    \multirow{3}{3cm}{Learning rate scheduler} & type & inverse\_sqrt \\
    & warmup updates & 400 \\
    & warmup init learning rate & 1e-7 \\
    \cmidrule(lr){1-1} \cmidrule(lr){2-2} \cmidrule(lr){3-3}
    \multirow{4}{3cm}{Training} & batch size & 2,048 tokens \\
    & tokens-per-sample & 128 tokens \\
    & sample-break-mode & none \\ 
    & epochs & 10 \\ \bottomrule
        \end{tabular}
        \subcaption{LSTM.}
        \label{tbl:hyperparam_lstm}
\end{minipage}

\begin{minipage}[t]{\hsize}
\renewcommand{\arraystretch}{0.4}
    \centering
    \begin{tabular}{p{3cm}p{5cm}p{4.5cm}} \toprule
     \multirow{8}{1cm}{Fairseq model} 
      & share-decoder-input-output-embed & True \\
      & embed\_dim & 64 \\
      & hiden\_size & 64 \\
      & layers & 2 \\
      & dropout & 0.1 \\
      & \#params. & 49K \\
    \cmidrule(lr){1-1} \cmidrule(lr){2-2} \cmidrule(lr){3-3}
    \multirow{5}{*}{Optimizer} & algorithm & AdamW \\
    & learning rates & 5e-4 \\
    & betas & (0.9, 0.98) \\
    & weight decay & 0.01 \\
    & clip norm & 0.0 \\
    \cmidrule(lr){1-1} \cmidrule(lr){2-2} \cmidrule(lr){3-3}
    \multirow{3}{3cm}{Learning rate scheduler} & type & inverse\_sqrt \\
    & warmup updates & 400 \\
    & warmup init learning rate & 1e-7 \\
    \cmidrule(lr){1-1} \cmidrule(lr){2-2} \cmidrule(lr){3-3}
    \multirow{4}{3cm}{Training} & batch size & 2,048 tokens \\
    & tokens-per-sample &  128 tokens \\
    & sample-break-mode & none \\ 
    & epochs & 10 \\ \bottomrule
        \end{tabular}
        \subcaption{RNN.}
        \label{tbl:hyperparam_rnn}
\end{minipage}
\caption{Hyperparameters of LMs.}
\label{tbl:hyper_params}
\end{table*}

\subsection{Dataset Generation}
The \textsc{Original}, \textsc{Short}, \textsc{Medium} and \textsc{Long} datasets are the same ones used by \citet{elnaggar2025wordordersfacilitatelength}. 

\begin{itemize}
    \item \textbf{\textsc{Short} Dataset:} The sentences in this dataset are created by generating all possible category combinations, which are referred to as templates, of length 3 to 8 words.
    The templates are then parsed and assigned to the ALs where they would produce a valid parse, i.e. result in an S node spanning the input.
    All lengths are represented equally in the dataset, and within each length, the different templates are also represented equally.
    \item \textbf{\textsc{Medium} Dataset:} Like the \textsc{Short} dataset, this dataset is created by generating all possible templates of lengths 9-10 words.
    The templates are then parsed and assigned to ALs where they produce a valid parse. 
    Similarly, all lengths are represented equally and within each length all templates are represented equally. 
    The lexicon is then randomly sampled to create unique sentences.
    \item \textbf{\textsc{Long} Dataset:} To generate the \textsc{Long} dataset, the templates for the \textsc{Short} and \textsc{Medium} datasets are combined in 3 ways, as shown in Figure~\ref{fig:examples_template_extension}:
    \begin{enumerate}
        \item Appended end to end 
        \item Concatenated with a conjunction (Fig.~\ref{fig:examples_template_extension_append_and}),
        \item Embedded with a conjunction (Fig.~\ref{fig:examples_template_extension_embed}).
    \end{enumerate}
    The resulting longer templates are parsed to filter out ungrammatical ones.
    Because there are millions of valid templates of length 11-20, 20K templates are randomly sampled, and for each one, the lexicon is sampled.
    It is worth noting that for the English-like language configuration \texttt{0101101}, the vast majority of the valid extended templates will have been created by concatenation with a conjunction (b) or embedding with a conjunction (c). 
    However, it may be possible for the end to end appending of templates to result in valid templates in other language configurations. 
\end{itemize}

\subsection{Dataset Statistics}
We show the occurrences of the different categories in the templates in \citet{elnaggar2025wordordersfacilitatelength} for lengths 3-10 in Figure~\ref{fig:template-avg-category-count-per-length}.
The occurrences of the templates used to create the \textsc{Long} dataset are shown in Figure~\ref{fig:template-avg-category-count-per-length-long-sampled}, and in Figure~\ref{fig:template-avg-category-count-per-length-long-all} for all generated templates of length 11-20.
We can see in Figure~\ref{fig:avg-category-count-per-length-short-long} that the number of categories that appear in shorter sentences is smaller than in longer sentences, and that the number of category occurrences increases with sentence length.
We show in Figure~\ref{fig:operations_count} the average number of different combinatory operations that are required to derive the templates of different lengths.

\section{Model Hyperparameters}\label{appendix:model_hyperparameters}
We use exactly the same model hyperparameters as \citet{elnaggar2025wordordersfacilitatelength}. 
These are summarised in Table~\ref{tbl:hyper_params}.

\end{document}